**Title**
Enabling faster and more reliable sonographic assessment of gestational age through machine learning

**Authors**
Chace Lee[1†]; Angelica Willis[1†]; Christina Chen[1]; Marcin Sieniek[1]; Akib Uddin[1]; Jonny Wong[1]; Rory Pilgrim[1]; Katherine Chou[1]; Daniel Tse[1]; Shravya Shetty[1‡]; Ryan G. Gomes[1‡]

**Affiliations**
[1]Google Health, Palo Alto, CA, USA

[†]Chace Lee and Angelica Willis contributed equally to this work.
[‡]Shravya Shetty and Ryan G. Gomes jointly supervised this work.

**Abstract**
Fetal ultrasounds are an essential part of prenatal care and can be used to estimate gestational age (GA). Accurate GA assessment is important for providing appropriate prenatal care throughout pregnancy and identifying complications such as fetal growth disorders. Since derivation of GA from manual fetal biometry measurements (head, abdomen, femur) are operator-dependent and time-consuming, there have been a number of research efforts focused on using artificial intelligence (AI) models to estimate GA using standard biometry images, but there is still room to improve the accuracy and reliability of these AI systems for widescale adoption. To improve GA estimates, without significant change to provider workflows, we leverage AI to interpret standard plane ultrasound images as well as "fly-to" ultrasound videos, which are 5-10s videos automatically recorded as part of the standard of care before the still image is captured. We developed and validated three AI models: an image model using standard plane images, a video model using fly-to videos, and an ensemble model (combining both image and video). All three were statistically superior to standard fetal biometry-based GA estimates derived by expert sonographers, the ensemble model has the lowest mean absolute error (MAE) compared to the clinical standard fetal biometry (mean difference: -1.51 ± 3.96 days, 95% CI [-1.9, -1.1]) on a test set that consisted of 404 participants. We showed that our models outperform standard biometry by a more substantial margin on fetuses that were small for GA. Our AI models have the potential to empower trained operators to estimate GA with higher accuracy while reducing the amount of time required and user variability in measurement acquisition.

**Introduction**

Fetal ultrasound is the cornerstone of prenatal imaging and can provide crucial information to guide maternal-fetal care, such as estimated gestational age (GA) and fetal growth disorders. Currently, the clinical standard of estimating GA and diagnosis of fetal growth disorders are determined through manual acquisition of fetal biometric measurements such as biparietal diameter, head circumference, abdominal circumference, femur length, and crown-rump length. Numerous regression formulae for GA estimation exist based on different combinations of fetal biometric measurements. Hadlock's formula is one of the most popular formulas and is included with most ultrasound equipment packages. Previous studies have suggested that while fetal biometric measurements were generally reproducible, there was increased variance later in pregnancy, which is when many key clinical decisions (such as delivery scheduling and medication administration if needed) are made.[1,2]

The accuracy and efficiency of biometric measurements is dependent on the skill and experience of the sonographer. Factors like fetal movement and difficult fetal positioning can make it difficult to accurately position the ultrasound probe to acquire biometry measurements. There has been extensive research on using artificial intelligence (AI) systems to assist in estimating GA, typically through automatic estimation of biometric parameters that are then used in the Hadlock formula.[3–7] We have recently shown that GA model estimation that directly used ultrasound videos of pre-defined sweeps was non-inferior to standard fetal biometry estimates.[8]

In this study, we further extend the use of ultrasound videos by developing three end-to-end AI models: a) image model - using fetal ultrasound images captured by sonographers during biometry measurements, b) video model - using "fly-to" videos, or 5 to 10 seconds of video immediately before image capture, c) ensemble model - using both images and fly-to videos. All data were collected retrospectively during standard biometry measurements. All three models directly estimate GA, without requiring measurement acquisition or use of regression formulae. To our knowledge, this is the first attempt on using AI on standard of care ultrasound videos to predict GA directly for all trimesters without estimating biometry measurements from standard plane images.

**Results**

Our models were developed and evaluated using datasets prospectively collected as part of the FAMLI study [9]. Our evaluation was performed on a test set consisting of patients independent of those used for AI development. The primary test set consisted of 407 women with standard of care ultrasound scans performed by expert sonographers at University of North Carolina (UNC) Healthcare, Chapel Hill, NC, USA and at community clinics in Lusaka, Zambia. Complete sets of ultrasound fetal biometry images and fly-to videos data collected with the SonoSite Turbo-M or Voluson S6 ultrasound machine were available for 404 of these 407 participants, corresponding to 677 study visits.

The disposition of test set study participants used in the following analysis are summarized in STARD diagrams (Extended Data Figure 1). The characteristics of study participants who are included in the test set analyses are shown in Extended Data Table 1. Among study visits conducted by sonographers, 63 (9.3%) women had at least one visit during the first trimester, 235 (34.7%) women had at least one visit during the second trimester, and 379 (56.0%) had one or more visits in the third trimester.

**AI gestational age estimation using ultrasound images and fly-to videos**
The primary analysis outcome for GA was the mean difference in absolute error between the GA model estimate and the clinical standard estimate, with the ground truth GA extrapolated from the initial GA estimated at the initial exam. The ground truth GA at each subsequent visit was calculated as: GA at initial exam plus number of days since baseline visit. Statistical estimates and comparisons were computed after randomly selecting one study visit per patient for each analysis group, to avoid combining correlated measurements from the same patient.

GA analysis results are summarized in Table 1. The overall MAE for the image GA model is lower compared to the MAE for the standard fetal biometry estimates (mean difference -1.13 ± 4.18 days, 95% CI [-1.5, -0.7]), the upper limit of the 95% CI for the difference in MAE values was negative, indicating statistical superiority of the model.

The overall MAE of the video model and ensemble models are significantly lower compared to the standard fetal biometry; the ensemble model has the lowest MAE (mean difference -1.51 ± 3.96 days, 95% CI [-1.9, -1.1]), followed by the video model (mean difference -1.48 ± 4.05 days, 95% CI [-1.9, -1.1]). For both models, the upper limit of the 95% CI for the difference in MAE values was negative, indicating statistical superiority.

In addition, secondary subgroup analyses were performed, with subgroups including pregnancy trimester (first, second or third), by country (US and Zambia), and device manufacturers (SonoSite Turbo-M or Voluson S6). For each subgroup analysis, estimates and comparisons were computed after randomly selecting one study visit per patient for patients eligible for the subgroups, to avoid combining correlated measurements from the same patient. Subgroup analysis for trimester, countries and devices are provided in Tables 2 and 3. The result shows that our models generalize well across countries, devices and second and third trimesters, with lower MAE compared to the standard fetal biometry estimates. The upper limit of the 95% CI for the difference in MAE values was less than 0.1 day, except for the first trimester, where the smaller sample size for the first trimester dataset size broadened confidence intervals. There was a trend towards increasing error for all models and standard fetal biometry procedures with gestational week.

**Gestational age estimation compared with alternative formulae**
The formulae derived in Hadlock et al.[2] are the standard of care in the United States and many other countries around the world; however, these formulae, having been derived based on a limited population of 361 middle-class Caucasian women in Texas, might not generalize as well as alternative formulae developed with broader, population-based data. For this reason, we compared the predictions on the FAMLI population with two additional formulae,

INTERGROWTH-21st and NICHD, that have shown promise as potential alternatives to Hadlock.[10,11]

GA analysis results are summarized in Table 4, the ensemble model has the lower MAE compared to NICHD (mean difference -1.23 ± 4.04 days, 95% CI [-1.6, -0.8]), and INTERGROWTH-21st (mean difference -2.69 ± 5.54 days, 95% CI [-3.3, -2.1]), the upper limits of the 95% CI for the difference in MAE values were less negative, indicating statistical superiority.

Analysis of per country performance is summarized in Extended Table 2, which shows that the accuracy of Hadlock-based GA estimates are close to Hadlock in the US (NICHD MAE: 4.79 ± 4.16, Hadlock MAE: 4.90 ± 4.32), while outperforming Hadlock significantly in the Zambia population (NICHD MAE: 4.96. ± 4.84, Hadlock MAE: 5.62 ± 5.2). The INTERGROWTH-21st formula performs significantly worse than Hadlock and NICHD across both populations. Our ensemble model estimate is compared against NICHD; the result shows the ensemble model has a lower MAE for both US (mean difference 3.58 ± 2.79 days, 95% CI [-1.9, -0.7]) and Zambia (mean difference 3.70 ± 3.57 days, 95% CI [-1.8, -0.6]), demonstrating robustness and statistical superiority on all subgroups.

**Performance of model on small for gestational age (SGA) and large for gestational age (LGA) cases**
Biometry-based methods of determining gestational age are predisposed to underestimation and overestimation of SGA and LGA fetuses, respectively. To understand if the AI model was affected, we performed an analysis of accuracy achieved on fetuses smaller or larger than expected for their GA based on fetal abdominal circumference (AC) measured for the given population and ground truth gestational week, using the 10th percentile threshold for SGA[12] and 90th percentile threshold for LGA.

The analysis of our model shows that it outperforms Hadlock by a wider margin on every SGA or LGA-sized subgroup of data, compared to cases in the "Normal" (non-SGA and non-LGA) patient set (Table 5). SGA performance (ensemble model: mean difference -3.46 ± 5.69, 95% CI [-5.0, -1.9]) compared with normally sized fetuses (ensemble mode: mean difference -1.08 ± 3.34, 95% CI -1.4, -0.7). The largest performance improvement over the Hadlock formula, though smallest in sample size (n = 26 patients), can be observed in severe SGA cases (ensemble model: mean difference -4.45 ± 6.96, 95% CI [-7.3, -1.6]), defined by a 3rd percentile AC threshold. We observed a similar phenomenon of more substantial performance gains for these traditionally challenging subgroups with still image models (see Extended Data Table 3).

**Discussion**
In this study, we demonstrated that our end-to-end AI models can provide GA estimates with improved precision. We show that our AI systems can use both images and fly-to videos, which are already collected as part of the standard fetal biometry measurements. Our three models (image, video, ensemble) each provide statistically superior GA estimates when compared to the clinical standard fetal biometry. Unlike previously described dating methods using anatomical measurements, our models make use of the entire image or video without needing the operator to accurately place calipers for precise measurements.

We show that our models performed well across different trimesters, devices, and populations from two different countries. GA estimation is known to be less accurate as pregnancy progresses since the correlation between GA and physical size of the fetus is less pronounced.[13] We found that in the third trimester, our model's accuracy advantage relative to the clinical standard fetal biometry increased. This is particularly important because accurate GA estimation in the third trimester is essential for managing complications and making appropriate clinical decisions such as the timing of delivery.

Fetal growth restriction is a significant complication in pregnancy. Worldwide, 60% of neonatal deaths are associated with low birth weight.[14] Our models had a significant increase in relative performance over fetal biometry when evaluating fetuses who were SGA. These subgroups are particularly challenging for fetal biometry estimation as the formulae rely on fetal size measurements. More accurately identifying fetuses that are SGA, as opposed to misclassifying them as having a lower GA, would help guide critical clinical care decisions such as scheduling the delivery, administering medications like steroids for lung maturation, and transferring the neonate to a higher level of care after delivery.[15,16]

One limitation of this study is that several of our subgroups had small sample sizes. Collecting additional ultrasounds in the first trimester and of SGA cases will help further validate our observations. Additionally, while this study included patients from two countries, it is important to validate on a more diverse population to confirm generalizability, as fetal growth patterns differ in different populations.[17] Also of note, our model has not been tested on multifetal gestations or fetuses with abnormal anatomy. Lastly, while we show that our models achieve statistical significance, a prospective study is needed to evaluate clinical impact.

In conclusion, we show that our image model, video model, and ensemble model provide statistically superior GA estimation when compared to the clinical standard fetal biometry. Our models had a significant increase in relative performance over fetal biometry in the third trimester and when evaluating fetuses who were SGA. Since our models are built on data collected during routine fetal ultrasounds, they have the potential of being incorporated seamlessly into the routine clinical workflow. Our AI models have the potential to empower trained operators to estimate GA with higher accuracy while reducing the amount of time required and user variability in measurement acquisition.

# Tables and figures

**Table 1: Gestational age estimation overall performance.** Mean absolute error (MAE) and mean error (ME) between GA estimated using the AI models and ground truth, and the MAE and ME between the GA estimated using the standard fetal biometry ultrasound procedure and ground truth. One visit by each participant eligible for each subgroup was selected at random. The video model is an ensemble of the I3d and LSTM video models. The ensemble model combines predictions from the two video models and the image model.

| No. patients: 404, Average gold standard GA ± sd (days): 192.9 ± 53.3 | | | | |
|---|---|---|---|---|
| Estimation Method | Standard fetal biometry estimates | Ensemble Model | Video Model | Image Model |
| ME ± sd (days) | -1.44 ± 6.82 | -0.45 ± 4.81 | -0.54 ± 4.84 | -0.29 ± 5.32 |
| MAE ± sd (days) | 5.11 ± 4.73 | 3.6 ± 3.23 | 3.63 ± 3.24 | 3.97 ± 3.54 |
| MAE difference compared to standard fetal biometry mean difference ± sd (days) | Reference | -1.51 ± 3.96 | -1.48 ± 4.05 | -1.13 ± 4.18 |
| 95% CI of difference (days) | Reference | -1.9, -1.1 | -1.9, -1.1 | -1.5, -0.7 |

**Table 2: Gestational age estimation subgroup analysis split by trimester.** Mean absolute error (MAE) and mean error (ME) between GA estimated using the AI models and ground truth, and the MAE and ME between the GA estimated using the standard fetal biometry ultrasound procedure and ground truth. One visit by each participant eligible for each subgroup was selected at random.

| First Trimester  No. patients: 55, Average gold standard GA ± sd (days): 77.1 ± 14.2 | | | | |
|---|---|---|---|---|
| Estimation Method | Standard fetal biometry estimates | Ensemble Model | Video Model | Image Model |
| ME ± sd (days) | 0.53 ± 3.12 | 1.22 ± 3.63 | 1.36 ± 3.72 | 0.92 ± 4.17 |
| MAE ± sd (days) | 2.35 ± 2.1 | 2.94 ± 2.43 | 3.17 ± 2.35 | 3.04 ± 2.98 |
| MAE difference compared to standard fetal biometry mean difference ± sd (days) | Reference | 0.6 ± 2.59 | 0.82 ± 2.58 | 0.69 ± 3.16 |
| 95% CI of difference (days) | Reference | -0.1, 1.3 | 0.1, 1.5 | -0.2, 1.5 |
| Second Trimester  No. patients: 171, Average gold standard GA ± sd (days): 154.7 ± 26.9 | | | | |
| ME ± sd (days) | -0.53 ± 4.2 | 0.67 ± 3.44 | 0.38 ± 3.61 | 1.23 ± 3.64 |
| MAE ± sd (days) | 3.42 ± 2.48 | 2.84 ± 2.05 | 2.93 ± 2.14 | 3.11 ± 2.25 |
| MAE difference compared to standard fetal biometry mean difference ± sd (days) | Reference | -0.58 ± 2.44 | -0.48 ± 2.56 | -0.31 ± 2.71 |
| 95% CI of difference (days) | Reference | -0.9, -0.2 | -0.9, -0.1 | -0.7, 0.1 |
| Third Trimester  No. patients: 291, Average gold standard GA ± sd (days): 225.8 ± 17.9 | | | | |
| ME ± sd (days) | -2.09 ± 7.55 | -0.81 ± 5.37 | -0.95 ± 5.32 | -0.53 ± 6.04 |
| MAE ± sd (days) | 6.0 ± 5.02 | 3.99 ± 3.68 | 3.98 ± 3.66 | 4.55 ± 4.0 |
| MAE difference compared to standard fetal biometry mean difference ± sd (days) | Reference | -2.01 ± 4.38 | -2.03 ± 4.41 | -1.46 ± 4.62 |
| 95% CI of difference (days) | Reference | -2.5, -1.5 | -2.5, -1.5 | -2.0, -0.9 |

**Table 3: Subgroup analysis split by country and manufacturer device.** We split the result into three subgroups (US-GE, Zambia-GE, and Zambia-Sonosite). SonoSite was only used in Zambia. Average gestational age is higher in Zambia vs the US due to a higher ratio of third trimester studies. Mean absolute error (MAE) and mean error (ME) between GA estimated using the AI models and ground truth, and the MAE and ME between the GA estimated using the standard fetal biometry ultrasound procedure and ground truth. One visit by each participant eligible for each subgroup was selected at random.

| US - GE  No. patients: 180, Average gold standard GA ± sd (days): 164.4 ± 60.4 | | | | |
|---|---|---|---|---|
| Estimation Method | Standard fetal biometry estimates | Ensemble Model | Video Model | Image Model |
| ME ± sd (days) | 1.14 ± 5.9 | 0.58 ± 4.4 | 0.51 ± 4.44 | 0.72 ± 4.96 |
| MAE ± sd (days) | 4.48 ± 4.0 | 3.49 ± 2.74 | 3.48 ± 2.79 | 3.9 ± 3.13 |
| MAE difference compared to standard fetal biometry mean difference ± sd (days) | Reference | -0.99 ± 3.44 | -0.99 ± 3.53 | -0.57 ± 3.68 |
| 95% CI of difference (days) | Reference | -1.5, -0.5 | -1.5, -0.5 | -1.1, -0.0 |
| Zambia - GE  No. patients: 77, Average gold standard GA ± sd (days): 216.8 ± 32.5 | | | | |
| ME ± sd (days) | -5.23 ± 5.7 | -2.66 ± 3.62 | -3.03 ± 3.66 | -1.92 ± 4.33 |
| MAE ± sd (days) | 6.01 ± 4.8 | 3.65 ± 2.6 | 3.81 ± 2.82 | 3.97 ± 2.57 |
| MAE difference compared to standard fetal biometry mean difference ± sd (days) | Reference | -2.36 ± 4.38 | -2.2 ± 4.52 | -2.05 ± 4.66 |
| 95% CI of difference (days) | Reference | -3.4, -1.4 | -3.2, -1.2 | -3.1, -1.0 |
| Zambia - Sonosite  No. patients: 176, Average gold standard GA ± sd (days): 212.5 ± 34.1 | | | | |
| ME ± sd (days) | -2.56 ± 6.8 | -0.68 ± 5.23 | -0.72 ± 5.14 | -0.61 ± 5.87 |
| MAE ± sd (days) | 5.12 ± 5.12 | 3.6 ± 3.84 | 3.58 ± 3.74 | 4.09 ± 4.24 |
| MAE difference compared to standard fetal biometry mean difference ± sd (days) | Reference | -1.52 ± 3.97 | -1.54 ± 4.08 | -1.03 ± 4.15 |
| 95% CI of difference (days) | Reference | -2.1, -0.9 | -2.2, -0.9 | -1.7, -0.4 |

**Table 4: Comparison against alternative biometry regression formulae.** We compared the mean absolute error (MAE) and mean error (ME) of our Video + Image Ensemble model against that of alternative fetal biometry-based regression formulae Intergrowth-21st and NICHD, in addition to Hadlock. Error Difference and 95% confidence interval (CI) denote the ensemble model MAE less the MAE for each fetal biometry-based procedure. We performed the comparison for second and third trimester cases, as the NICHD formula and the Intergrowth-21st formula are not applicable to first trimester cases.

| Overall (Second+Third Trimester) No. patients: 379, Average gold standard GA ± sd (days): 202.8 ± 41.7 | | | | |
|---|---|---|---|---|
| Estimation Method | Hadlock | Intergrowth-21st | NICHD | Ensemble Model |
| ME ± sd (days) | 1.61 ± 7.04 | 4.22 ± 7.31 | 0.23 ± 6.69 | -0.47 ± 4.88 |
| MAE ± sd (days) | 5.32 ± 4.87 | 6.35 ± 5.36 | 4.89 ± 4.57 | 3.65 ± 3.27 |
| MAE difference compared to ensemble model mean difference ± sd (days) | -1.68 ± 4.1 | -2.69 ± 5.54 | -1.23 ± 4.04 | Reference |
| 95% CI of difference (days) | -2.1, -1.3 | -3.3, -2.1 | -1.6, -0.8 | Reference |

**Table 5: Comparison for fetuses that are small or large for their gestational age (SGA and LGA) based on Abdominal Circumference (AC).** We compare mean absolute error (MAE) and mean error (ME) of our Video + Image Ensemble Model against that of the standard of care Hadlock procedure for SGA and LGA fetuses. Small fetuses are defined by having an AC below the 10th percentile and are considered very small if the AC is below the 3rd percentile. Large fetuses are defined by having an AC above the 90th percentile for their gestational age. *Normal size for gestational age* is defined as not suspected of being SGA or LGA, while *SGA or LGA* represents cases in which either SGA or LGA is suspected. Two sets of AC percentiles were derived from the FAMLI dataset based on both US and Zambia populations.

| Subgroup / estimated SGA/LGA status | Ensemble Model Error ± sd (days) | | Hadlock Error ±sd (days) | | MAE difference compared to ensemble model mean difference ± sd (days) | 95% confidence interval | Number of unique patients |
|---|---|---|---|---|---|---|---|
| | MAE | ME | MAE | ME | | | |
| Overall | 3.53 ± 3.2 | -0.53 ± 4.74 | 4.8 ± 4.38 | -1.32 ±6.37 | -1.27 ± 3.7 | -1.6, -0.9 | 379 |
| SGA (AC percentile < 10) | 5.12 ± 5.72 | -4.41 ± 6.30 | 8.58 ± 7.46 | -7.88 ± 8.21 | -3.46 ± 5.69 | -5.0, -1.9 | 57 |
| severe SGA (AC percentile < 3) | 6.13 ± 7.99 | -5.46 ± 8.48 | 10.58 ± 9.65 | -10.04 ± 10.23 | -4.45 ± 6.96 | -7.3, -1.6 | 26 |
| LGA (AC percentile > 90) | 4.11 ± 2.86 | -0.85 ± 4.20 | 4.96 ± 4.01 | 2.60 ± 5.86 | -1.30 ± 3.62 | -2.0, 0.3 | 55 |
| Normal size for GA (10 < AC percentile < 90) | 3.28 ± 2.54 | -0.52 ± 4.12 | 4.36 ± 3.45 | -1.19 ± 5.44 | -1.08 ± 3.34 | -1.4, -0.7 | 327 |
| SGA or LGA (AC percentile < 10 OR > 90) | 4.61 ± 4.63 | -1.08 ± 6.46 | 6.83 ± 6.34 | -2.74 ± 8.93 | -2.22 ± 5.22 | -3.2, -1.2 | 108 |

**Figure 1: Gestational age estimation. Top Left:** Model and standard fetal biometry estimates mean absolute error (MAE) versus ground truth GA (four-week GA windows). **Top Right:** Model and standard fetal biometry estimate absolute error distribution. estimate more as gestational age increases). **Bottom Left:** Error distributions for ensemble model and standard fetal biometry procedure. **Bottom Right:** Paired errors for ensemble model and standard fetal biometry estimates in the same study visit. The errors of the two methods exhibit correlation, but the worst-case errors for the ensemble model have a lower magnitude than the standard fetal biometry method.

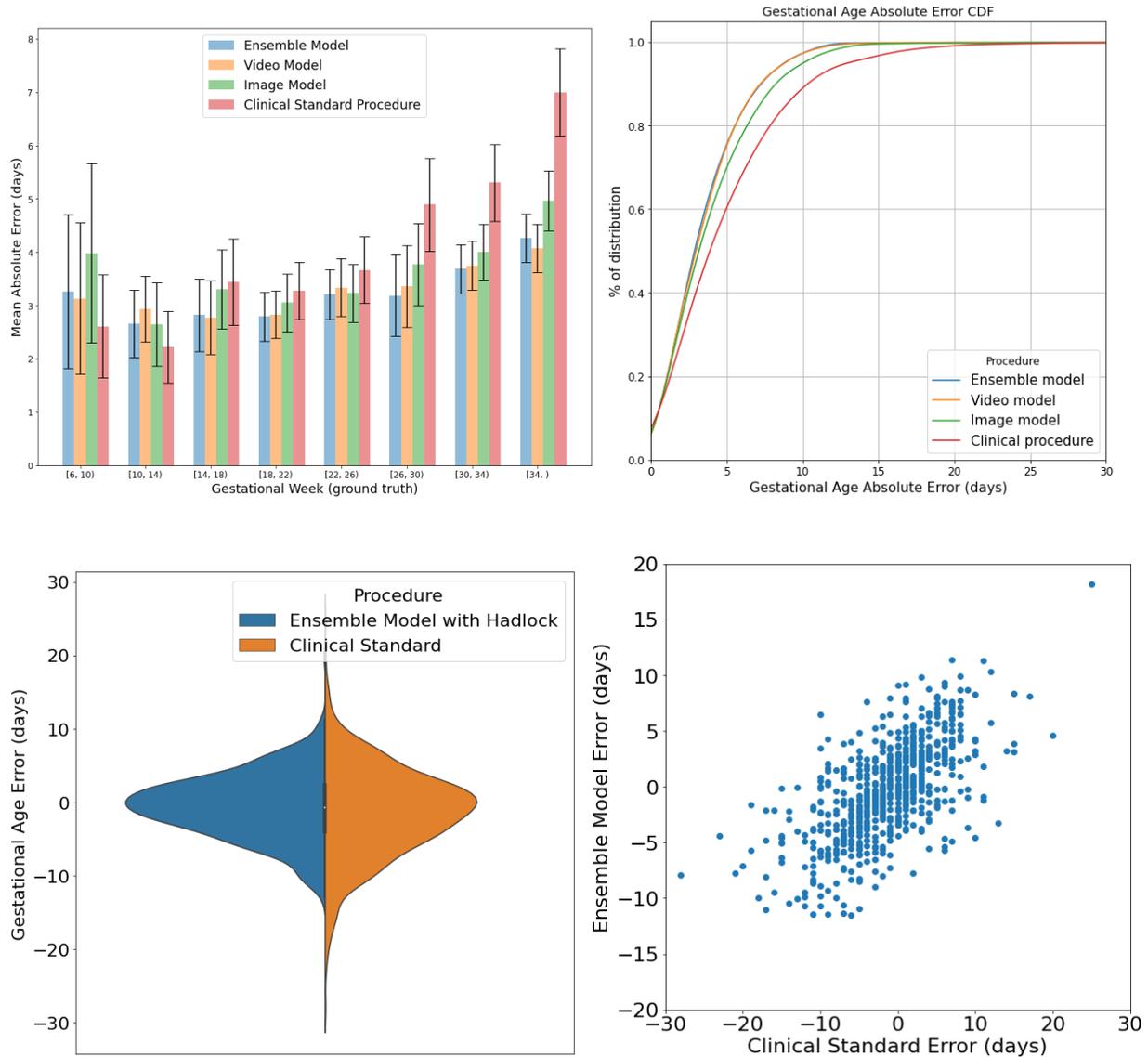

**Figure 2: Gestational age estimation comparison for fetuses that are small for their gestational age (SGA) based on abdominal circumference (AC).** Video + Image ensemble model and standard fetal biometry estimates mean absolute error (MAE) versus ground truth GA (four-week GA windows) on SGA fetuses (Top) and Severe SGA fetuses (Bottom). Small fetuses are defined by having an AC below the 10th percentile and are considered very small if the AC is below the 3rd percentile. P-values are calculated for each four-week GA window for the one-sided test with null hypothesis that the median of MAE differences (model MAE - standard fetal biometry estimates MAE) is positive.

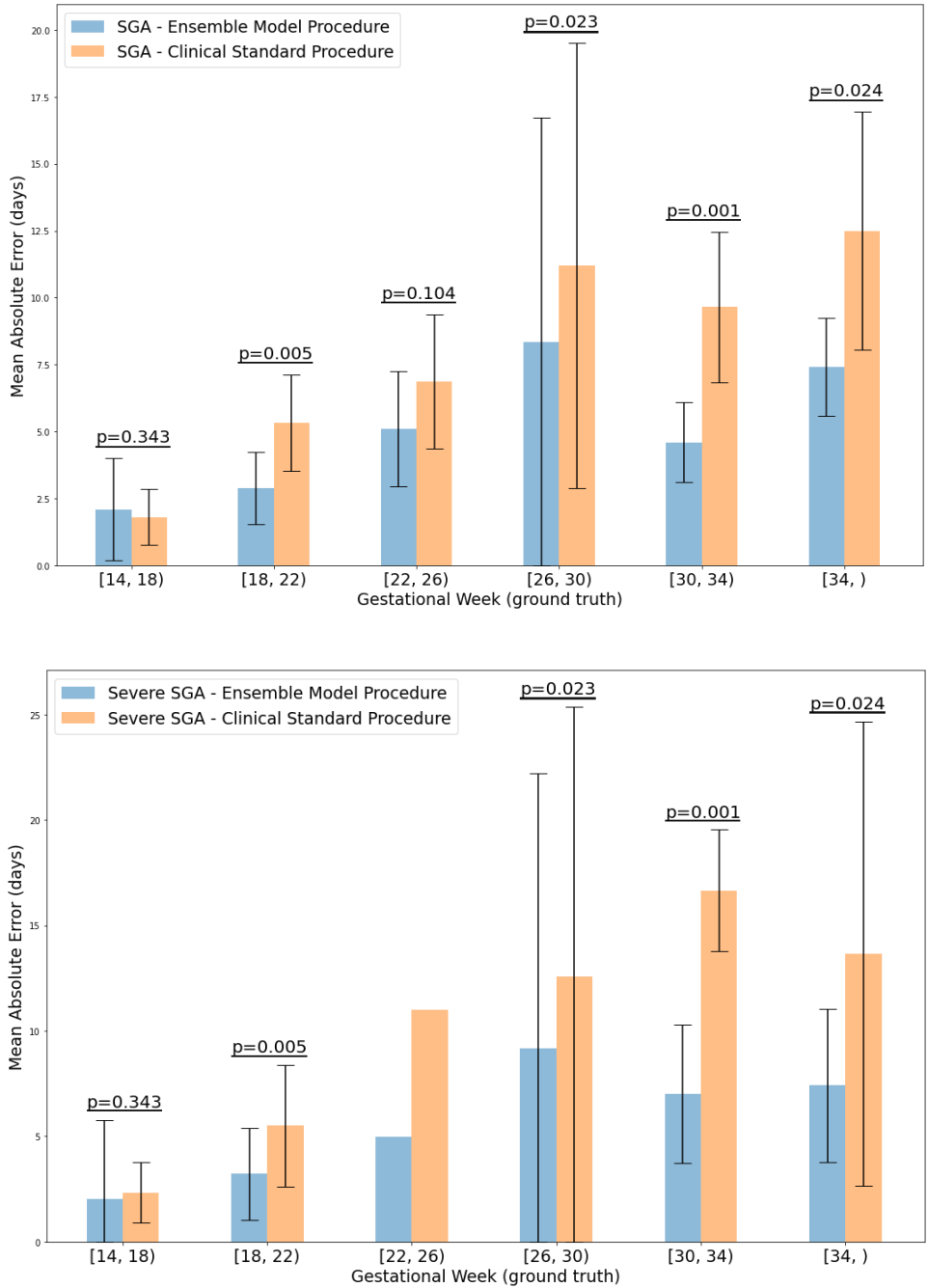

**Competing Interests and Funding**
This study was partially funded by Google LLC. Authors affiliated with Google are employees and own stock as part of the standard employee compensation package. This study was partially funded by the Bill and Melinda Gates Foundation (OPP1191684, INV003266). The conclusions and opinions expressed in this article are those of the authors and do not necessarily reflect those of the Bill and Melinda Gates Foundation.

**Acknowledgments**
We would like to thank Yun Liu for helpful feedback on the manuscript.

## Methods

**Algorithm development**

We developed three deep learning neural network models to predict GA: image, video, and an ensemble approach combining both. The image model generates GA prediction, measured in days, directly from each standard plane fetal biometry image, with image pixels of fixed dimension as input. The video model generated GA prediction directly from fly-to video sequences, with fixed-length sequences of image pixel values as input. The GA model additionally provided an estimate of its confidence in the estimate for a given video sequence or image. No intermediate fetal biometric measurements were required during training or generated during inference.

Our models were developed and evaluated using datasets prospectively collected as part of the Fetal Age Machine Learning Initiative (FAMLI)[9], which collected ultrasound data from study sites at Chapel Hill, NC (USA) and Lusaka, Zambia. All study participants provided written informed consent, and the research was approved by the UNC institutional review board and the biomedical research ethics committee at the University of Zambia. The goal of this prospectively collected dataset was to accelerate the development of technology to estimate gestational age. We trained our models on studies from all trimesters with images and videos captured by trained sonographers using standard ultrasound devices (SonoSite Turbo-M or Voluson S6), excluding images and videos captured using low-cost portable ultrasound device (ButterflyIQ) and novice studies, in an effort to enable applicability to device types used in standard of care. Study participants were assigned at random to one of 3 dataset splits: train, tune, or test. We used the following proportions: 60% train / 20% tune / 20% test. The tuning set was used for tuning model hyperparameters.

The image model was trained on all CRL standard plane images from first trimester studies and HC, AC and FL standard plane images from second and third trimester studies. The video model was trained on data that could be acquired in current stand-of-care procedures. Specifically, this included sonographer-acquired blind sweeps[18] (up to 15 sweeps per patient) as well as sonographer-acquired "fly-to" videos that capture five to ten seconds before the sonographer has acquired standard fetal biometry images. In the FAMLI study, GE machines (Voluson 8 in North Carolina; Logiq C3 in Zambia) were configured to export five to ten seconds of cine capture before the sonographer freezes/saves an image.

Cases consisted of multiple fly-to videos and standard biometry images, and our models generated predictions independently for each video sequence or image within the case. For the GA video model, each fly-to video was divided into multiple video sequences, We then aggregated the predictions to generate a single case-level estimate for GA (more details described below).

**Model architecture**

We developed two deep learning neural network models to predict GA from standard biometry images and fly-to videos. The GA regression models used the Best Obstetric Estimate of GA associated with the case as the training label for all video clips within the case. We defined gestational age prediction as a regression problem in which both image and video models produce an estimate of gestational age, measured in days.

Our image model was a single instance model trained on all standard plane biometry images for four standard anatomy types: crown-rump, fetal head, abdomen, and femur. The model architecture is depicted in Extended Data Figure 4. An independent gestational age and variance score was predicted for each image, and gestational age predictions were aggregated by the inverse weight of their predicted variance score to produce a final gestational age. This model uses an EfficientNet-B7[19] feature extractor and the final output layer feature maps were spatially average-pooled and transformed via a fully connected layer before final sigmoid units to compute the model's GA predictions for each image.

Our video model was assembled from an I3D convolutional model[20] and a convolutional recurrent model proposed for blindsweep GA prediction[8] which used a MobileNetV2[21] feature extractor. The model architecture is depicted in Extended Data Figure 5. The Inception I3D model we used was an I3D model based on Inceptionv1, this model did not perform temporal pooling in the first two max pooling layers; instead, it used 1 x 3 x 3 kernels and stride 1 in the time dimension. It used an Inception-v1 feature extractor applied to each video frame window volume with fixed sequence length (24 sampled frames), and the final output layer feature maps were spatially average-pooled and transformed via a fully connected layer before using the final softplus units to compute the model's GA predictions for each input frame window volume.

The I3D network and image model used a mean-variance regression loss function[22,23], which provided an estimate of expected variance by an additional softplus model output. The gestational age predictions were aggregated (ensembled) using the inverse weight of their predicted variance score to produce a final gestational age. The video model operated on log-transformed labels and used linear output units with offset -5.2 and label multiplier 3.43. To make predictions, we exponentiated the raw model output to recover the original scale. For the image model, we only used a linear label multiplier of 0.01.

**Data preprocessing**
For the video models, we used ½ temporal subsampling and clip length of 24 frames. Bilinear interpolation was used to resize the number of pixels in each video frame and image to a fixed scale according to the procedure described in the blindsweep GA prediction paper[8]. Video sequences were divided into multiple equal-length clips that corresponded to a fixed LSTM volume as aggregating predictions from multiple short clips works well for gestational age estimation.

The same image rescaling procedure was performed as a pre-processing step during model inference. The gestational age video model benefits from high-resolution images, and we used 576 x 432 pixels with α=0.0333 centimeters/pixel. The image model operates with lower resolution, and we used 320 x 240 pixels with α=0.06 centimeters/pixel.

**Training**
The image model was trained with multiple data augmentation strategies: random horizontal image flipping, random cropping, random rotation (with a maximum angle of 45.0 degrees), randomized saturation (with a range of [0.38,1.4]), brightness (with a maximum delta of 0.52), contrast (with a range of [0.34, 1.35]) and hue (with a maximum delta of 0.13). Model parameters were learned via AdamW optimization[24] with a batch size of 8. Slight dropout was also applied with a keep probability of 0.985. We use exponential decay earning rates with an initial learning rate of 4.56e-05, decay steps of 15490, and decay factor of 0.933.

The I3D model and convolutional recurrent model were trained with the same data augmentation procedure as described in the blindsweep GA prediction paper[8]. We applied dropout with a keep probability 0.8. Model parameters were learned via AdamW optimization[24] with a batch size of 8. The learning rate was decreased from an initial starting point according to a linear ramp-up schedule based on the current training step. The gestational age model began with a learning rate of 4.58e-04 and had a final rate of 4.58e-07, with training ending after 100k training steps. Dropout probabilities and learning rate schedules were optimized based on tuning set performance.

**Ensemble**
To improve model accuracy, we explored different ensemble configurations, ultimately selecting an ensemble of the still image model with the video models described above. Case GA predictions from the still image model, I3D model, and convolutional recurrent model were averaged after the inverse variance weighting procedure to give the final case prediction.

**Inference and confidence estimate**
Each case contains multiple fly-to videos of varying lengths and multiple standard plane images. We transformed each video into a set of overlapping video clips, where each clip contained a 24-frame sequence length and different starting frames. Both video and image models predict variance as a confidence estimate, which was used to aggregate predictions to generate case-level mean gestational age from different clips or images within the same case by inverse variance weighting as described in blindsweep GA model paper[8]. From our observations, clips with high confidence showed a clear anatomy view useful for gestational age prediction, while low confidence examples showed less informative anatomic views. The highest and lowest confidence clip examples from the fly-to video for each type of standard biometry are shown in the Extended Data Figure 2.

**Small-for-gestational-age and large-for-gestational-age**
We utilized the ISUOG definition of SGA[12] based on fetal abdominal circumference measured below the 10th percentile for the given population and ground truth gestational week. Similarly, a fetus is considered LGA if the abdominal circumference is measured as being greater than 90% of measurements within the same population, for their ground truth gestational week. We established these distinct percentile thresholds for each gestational week for each population (US and Zambia) across all patients of the dataset (train, tune, and test). To ensure the robustness of the derived SGA/LGA percentiles, we considered only gestations between 14 and 36 weeks inclusively, where the number of studies per week was at minimum 14, with an average of 51 cases per gestational week.

**Extended data figure 1: STARD diagrams. Standard procedure performed by trained sonographers (FAMLI study.)**

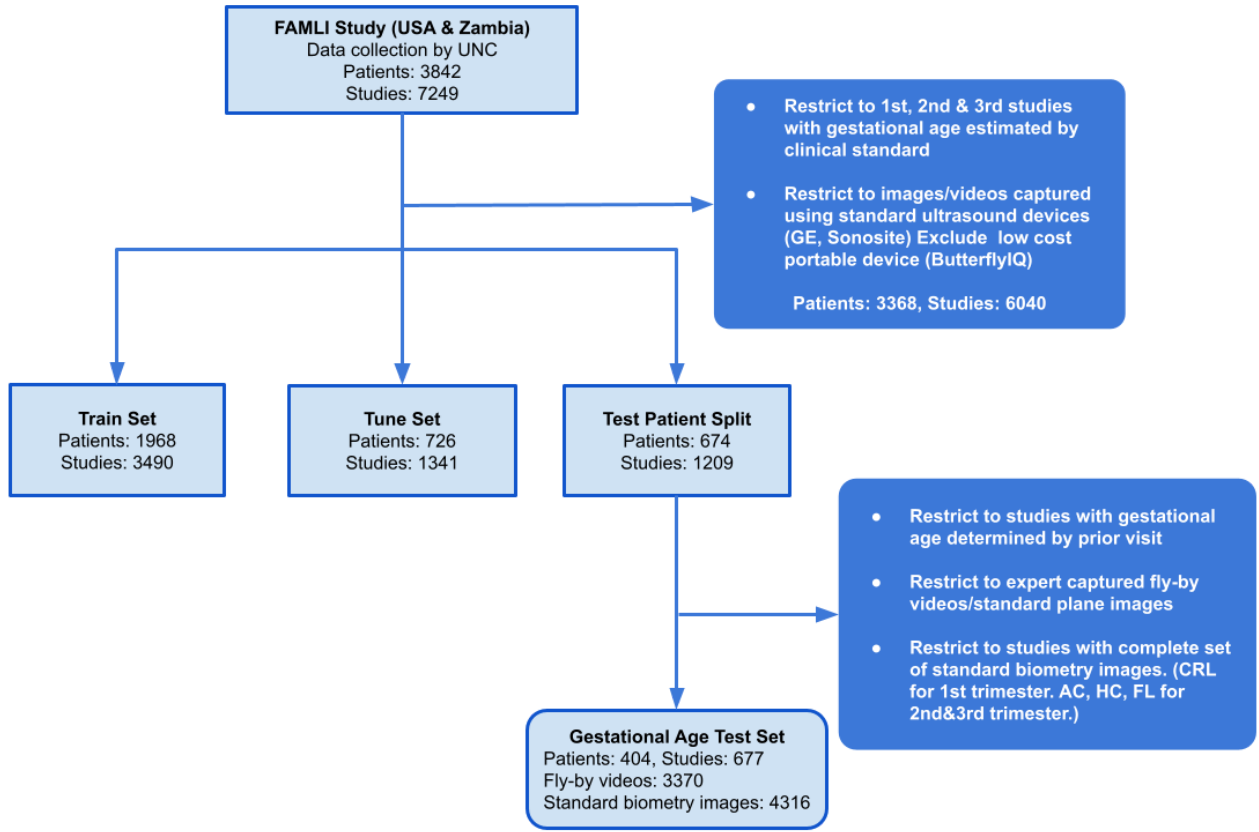

**Extended data figure 2: Fly-to video example images, the displayed frames are the central frame in a video sequence.** Left 2: Example fly-to video frames from clips with high confidence estimate for gestational age video model. Right 2: Example fly-to video frames from clips with low confidence estimate for gestational age video model. A, Example frames from head fly-to video. B, Example frames from abdomen fly-to video. C, Example frames from femur fly-to video. D, Example frames from crown-rump fly-to video.

A

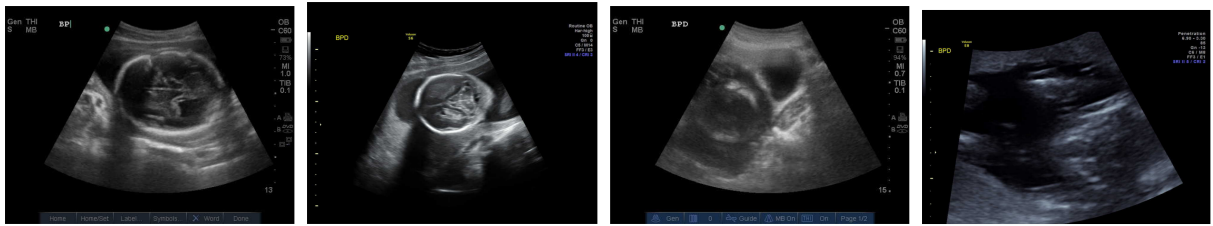

B

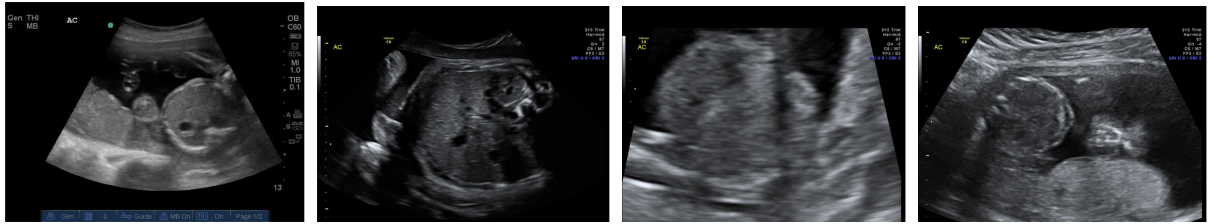

C

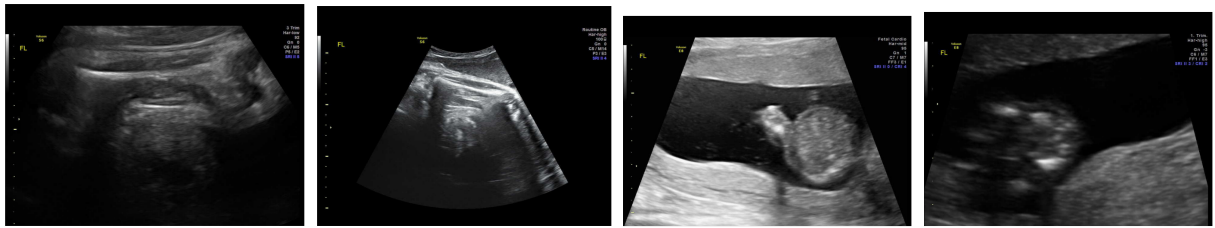

D

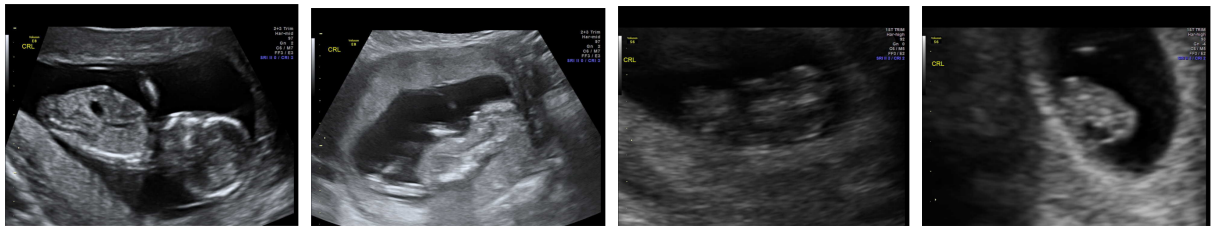

**Extended data figure 3: Image model architecture.** The EfficientNet network is used as a feature extractor applied to each image frame.

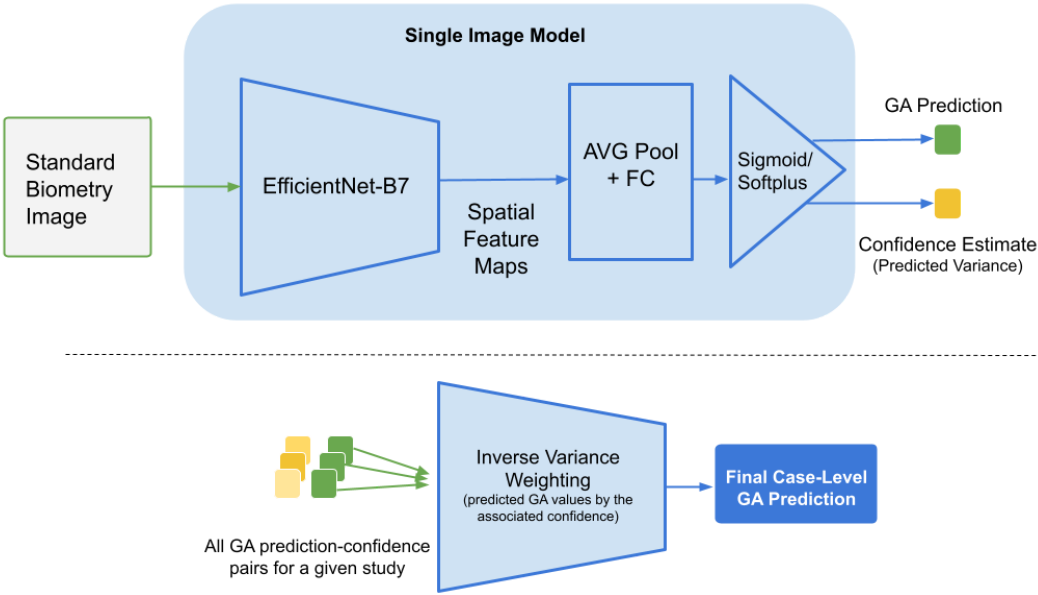

**Extended data figure 4: Video model architecture.** The video model is an ensemble of an I3D video model and an LSTM video model. For the I3D model, the Inception-I3D network is used as a feature extractor applied to video volume. For the LSTM video model, the MobileNetV2 network is used as a feature extractor applied to each video frame. The final feature layer of MobileNetV2 generates a sequence of image embeddings which is then processed by the grouped convolutional LSTM cell.

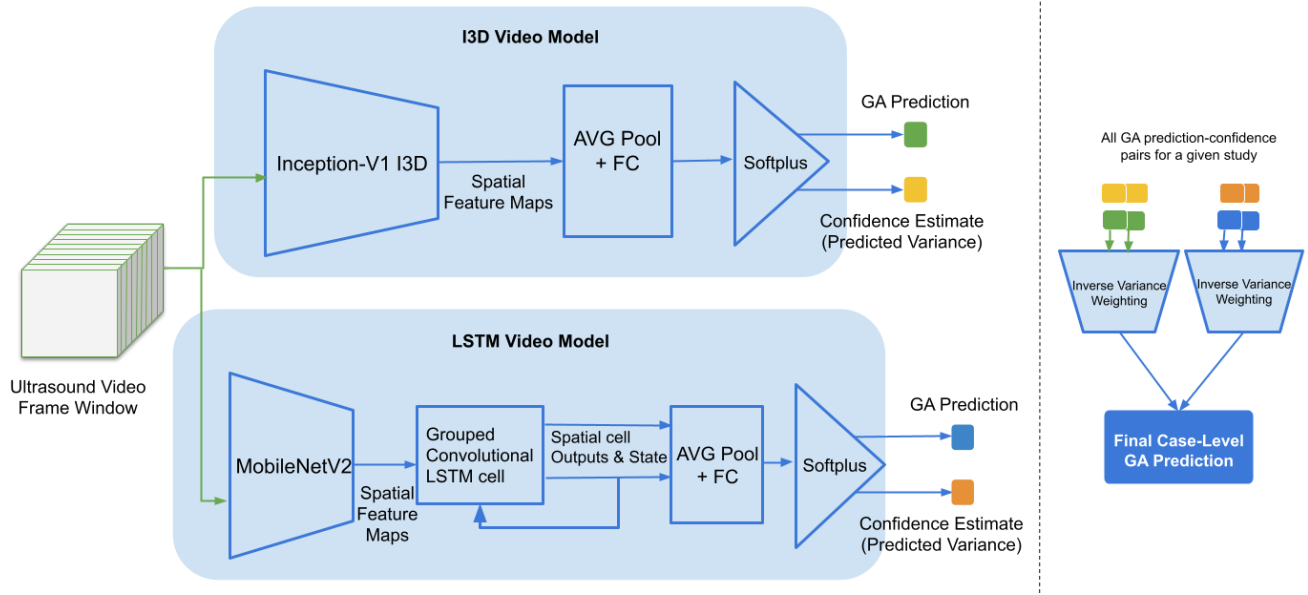

**Extended data figure 5: Model and standard fetal biometry estimate absolute error cumulative distribution plots for first trimester (Top Left), second trimester (Top Right), and third trimester (Bottom).**

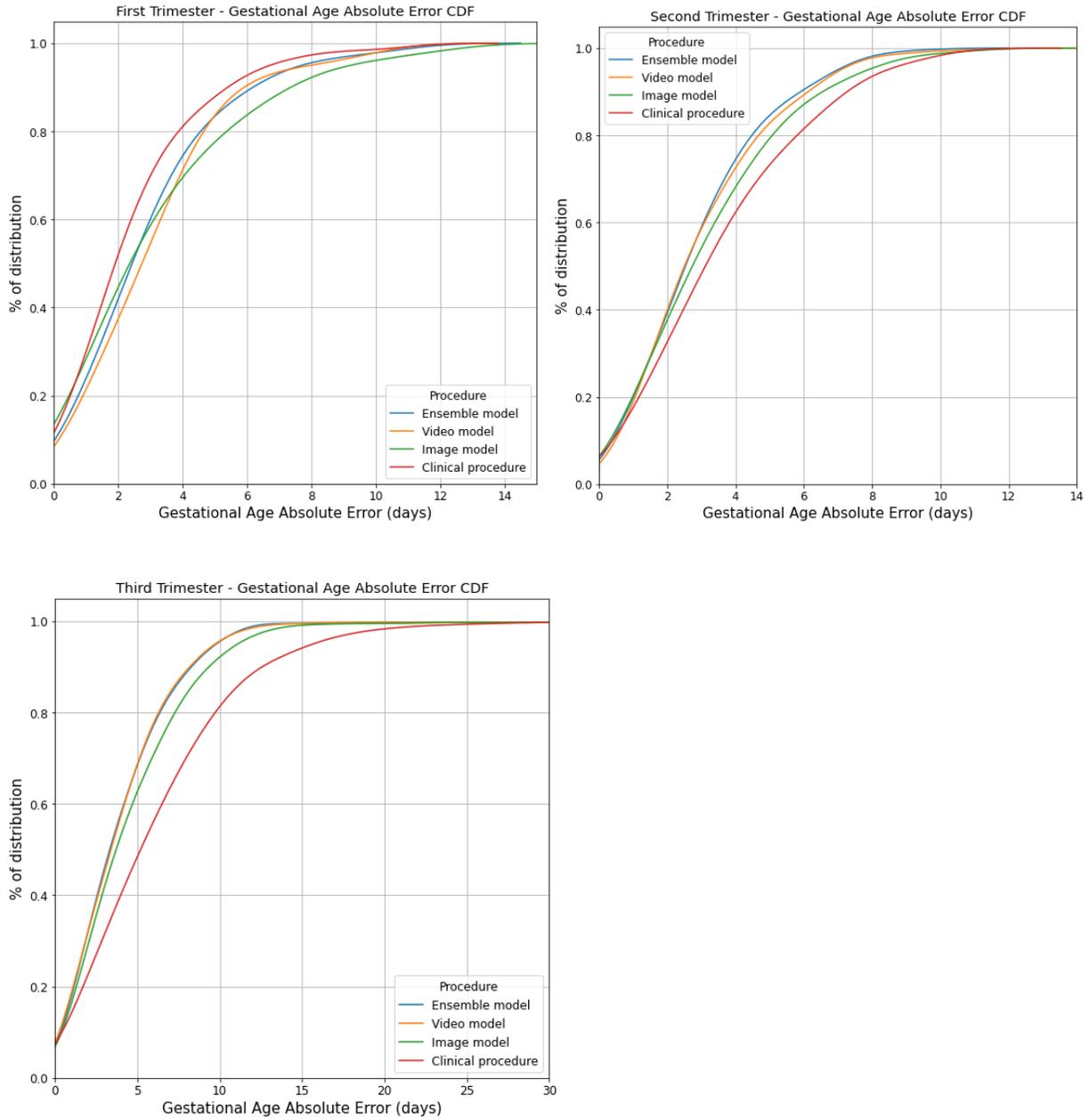

**Extended data figure 6: Gestational Age Mean Absolute Error versus Patient Characteristics.** MAE versus maternal age, height, and body mass index. Maternal height and Body Mass Index were only available for study participants in Zambia.

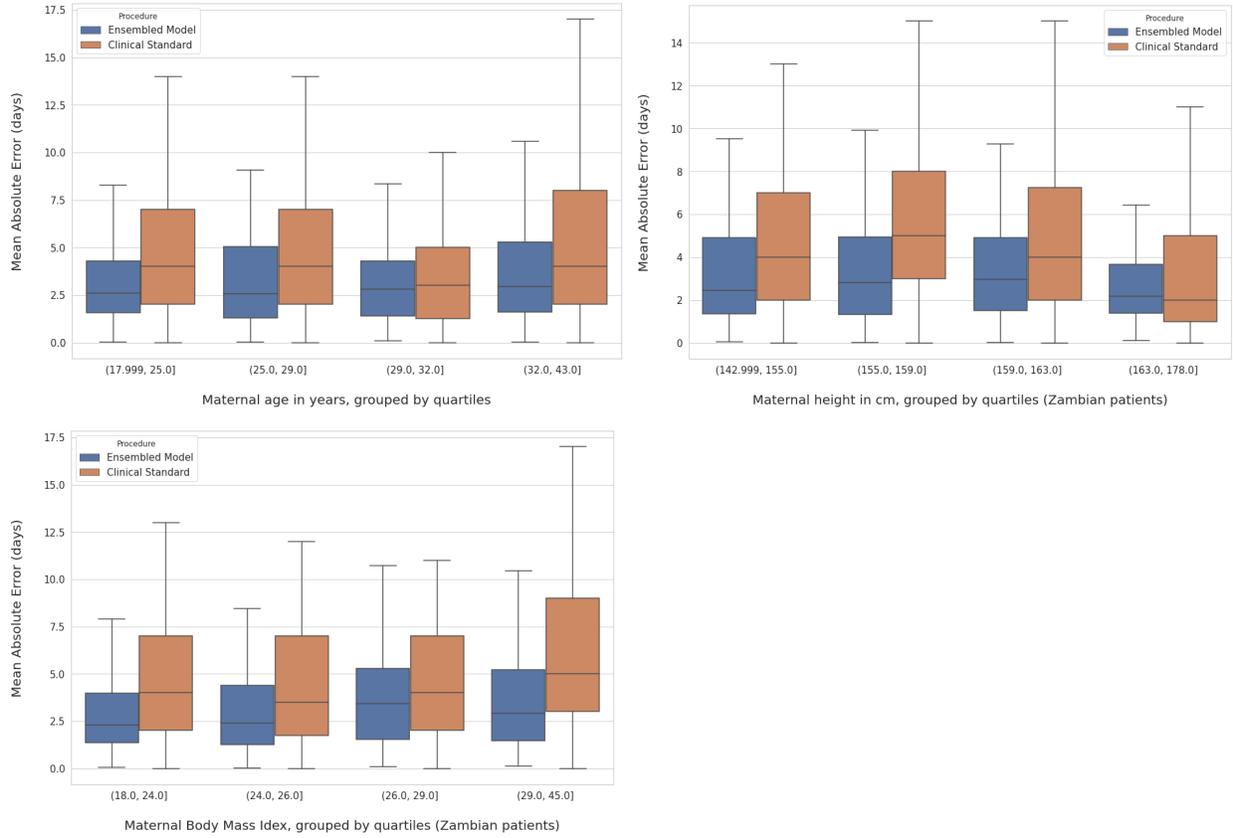

**Extended data figure 7: Fetal biometry ultrasound standard plane images for fetal biometry measurement. Top Left:** Abdominal circumference measurement. **Top Right:** Head circumference measurement. **Bottom Left:** Femur length measurement. **Bottom Right:** Crown-rump length measurement.

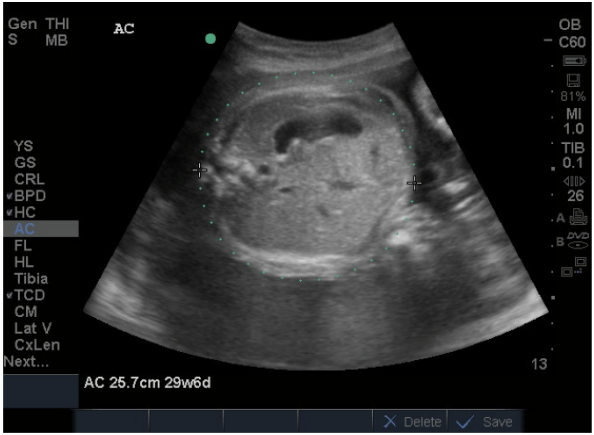
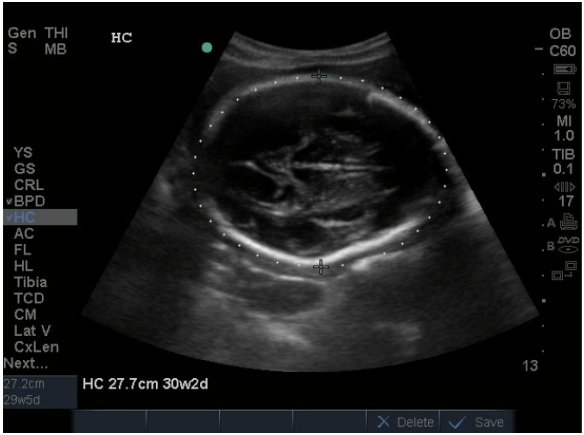
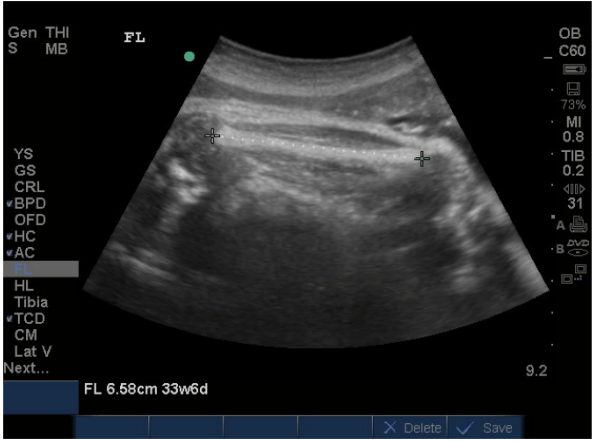
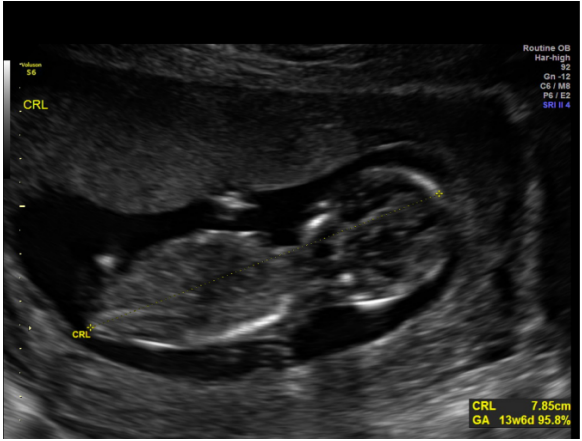

**Extended data table 1: Characteristics of study participants.**

| Variable | Sonographer group (n=404) |
|---|---|
| Participant age at enrollment (range) | 28.8 $\pm$ 5.6 years |
| Gestational age at first study visit | 17.6 $\pm$ 59.1 days |
| Total number of study visits during First trimester (%) | 63 (9.3%) |
| Total number of study visits during second trimester (%) | 235 (34.7%) |
| Total number of study visits during third trimester | 379 (56.0%) |

**Extended data table 2: Comparison against alternative biometry regression formulae split by trimester and country.** We compared the mean absolute error (MAE) and mean error (ME) of our Video + Image Ensemble model against that of alternative fetal biometry-based regression formulae Intergrowth-21st and NICHD, in addition to Hadlock. Error difference and 95% confidence interval denote the ensemble model MAE less the MAE for each fetal biometry-based procedure.

| Second Trimester<br>No. patients: 170, Average gold standard GA ± sd (days): 155.0 ± 26.8 | | | | |
|---|---|---|---|---|
| Estimation Method | Hadlock | Intergrowth-21st | NICHD | Ensemble Model |
| ME ± sd (days) | -0.52 ± 4.21 | 2.27 ± 4.56 | 0.47 ± 4.01 | 0.65 ± 3.44 |
| MAE ± sd (days) | 3.42 ± 2.48 | 3.93 ± 3.24 | 3.18 ± 2.47 | 2.83 ± 2.06 |
| MAE difference compared to ensemble model mean difference ± sd (days) | -0.59 ± 2.44 | -1.02 ± 3.49 | -0.26 ± 2.43 | Reference |
| 95% CI of difference (days) | -1.0, -0.2 | -1.5, -0.5 | -0.6, -0.1 | Reference |
| Third Trimester<br>No. patients: 291, Average gold standard GA ± sd (days): 225.8 ± 17.9 | | | | |
| ME ± sd (days) | -2.09 ± 7.55 | 5.34 ± 7.84 | 0.15 ± 7.21 | -0.82 ± 5.38 |
| MAE ± sd (days) | 6.0 ± 5.02 | 7.50 ± 5.80 | 5.54 ± 4.61 | 4.00 ± 3.67 |
| MAE difference compared to ensemble model mean difference ± sd (days) | -2.00 ± 4.38 | -3.51 ± 6.04 | -1.55 ± 4.24 | Reference |
| 95% CI of difference (days) | -2.5, -1.5 | -4.2, -2.8 | -2.0, -1.1 | Reference |
| US (Second and Third Trimester)<br>Number: 155, Average gold standard GA ± sd (days): 184.1 ± 46.8 | | | | |
| ME ± sd (days) | 1.14 ± 6.45 | 4.29 ± 7.34 | 2.19 ± 5.96 | 0.72 ± 4.49 |
| MAE ± sd (days) | 4.90 ± 4.32 | 6.20. ± 5.80 | 4.79. ± 4.16 | 3.58 ± 2.79 |
| MAE difference compared to ensemble model mean difference ± sd (days) | -1.32 ± 3.77 | -2.69 ± 5.61 | -1.28 ± 3.65 | Reference |
| MAE difference 95% CI (days) | -1.9, -0.7 | -3.6, -1.8 | -1.9, -0.7 | Reference |
| Zambia (Second and Third Trimester)<br>No. patients: 224, Average gold Average gold standard GA ± sd (days): 215.8 ± 31.9 | | | | |
| ME ± sd (days) | -3.52 ± 6.81 | 4.17 ± 7.31 | -1.13 ± 6.84 | -1.30 ± 4.98 |
| MAE ± sd (days) | 5.62 ± 5.20 | 6.45 ± 5.39 | 4.96. ± 4.84 | 3.70 ± 3.57 |
| MAE difference compared to ensemble model mean difference ± sd (days) | -1.92 ± 4.31 | -2.68 ± 5.50 | -1.19 ± 4.31 | Reference |
| 95% CI of difference (days) | -2.5, -1.4 | -3.4, -2.0 | -1.8, -0.6 | Reference |

**Extended data table 3: Still image model comparison for fetuses that are small (SGA) or large (LGA) for their gestational age based on abdominal circumference (AC).** We compare mean absolute error (MAE) and mean error (ME) of our Image Model against that of the standard fetal biometry Hadlock procedure for SGA and LGA fetuses. Small fetuses are defined by having an AC in the 10th percentile, and are considered very small if the AC is in the 3rd percentile. Large fetuses are defined by having an AC above the 90th percentile for their gestational age. *Normal size for gestational age* is defined as not suspected of being SGA or LGA, while *SGA or LGA* represents cases in which either SGA or LGA is suspected. Two sets of AC percentiles were derived from the FAMLI dataset based on both US and Zambia populations.

| Subgroup / estimated SGA/LGA status | Image Model Error ± sd (days) | | Hadlock Error ± sd (days) | | MAE difference compared to ensemble model mean difference ± sd (days) | 95% confidence interval | Number of unique patients |
|---|---|---|---|---|---|---|---|
| | MAE | ME | MAE | ME | | | |
| Overall | 3.94 ± 3.49 | -0.01 ± 5.28 | 4.8 ± 4.38 | -1.32 ± 6.37 | -0.85 ± 3.89 | -1.2, -0.5 | 379 |
| **SGA** (AC percentile < 10) | 5.41 ± 5.85 | -4.42 ± 6.65 | 8.58 ± 7.46 | -7.88 ± 8.21 | -3.16 ± 5.29 | -4.6, -1.8 | 57 |
| severe SGA (AC percentile < 3) | 6.76 ± 7.59 | -6.14 ± 8.12 | 10.58 ± 9.65 | -10.04 ± 10.23 | -3.81 ± 6.27 | -6.3, -1.3 | 26 |
| LGA (AC percentile > 90) | 4.63 ± 2.81 | 2.07 ± 5.03 | 4.96 ± 4.01 | 2.60 ± 5.86 | -0.34 ± 4.27 | -1.5, 0.8 | 55 |
| **Normal size for GA** (10 < AC percentile < 90) | 3.63 ± 2.89 | 0.22 ± 4.64 | 4.36 ± 3.45 | -1.19 ± 5.44 | -0.72 ± 3.57 | -1.1, -0.3 | 327 |
| SGA or LGA (AC percentile < 10 OR > 90) | 5.09 ± 4.66 | -1.29 ± 6.79 | 6.83 ± 6.34 | -2.74 ± 8.93 | -1.75 ± 5.08 | -2.7, -0.8 | 108 |